\documentclass{article}

\PassOptionsToPackage{numbers, square}{natbib}
\usepackage{arxiv}
\usepackage[numbers]{natbib}

\usepackage{subcaption}
\usepackage{tcolorbox}
\tcbuselibrary{breakable}
\usepackage{xcolor}
\usepackage{colortbl}
\usepackage{booktabs}
\usepackage{graphicx}
\usepackage{amsmath}
\usepackage{hyperref}
\usepackage{url}
\usepackage{float}
\usepackage{seqsplit}

\newcommand{\paragraphbf}[1]{\paragraph{\textbf{#1}}}
\definecolor{rowhl}{RGB}{232,245,233}
\definecolor{colhl}{RGB}{227,242,253}

\title{Data Turnstile: A Scalable Open Framework for \\ Function-Calling Data Generation}
\renewcommand{\shorttitle}{Data Turnstile: A Scalable Open Framework for Function-Calling Data Generation}

\author{
  Goutham Ramakrishnan\thanks{Equal contribution} \\
  Amazon AGI\\
  Bengaluru, India \\
  \texttt{gorama@amazon.com} \\
  \And
  Megha Sharma\footnotemark[1] \\
  Amazon AGI\\
  Bengaluru, India \\
  \texttt{meghshar@amazon.com} \\
}

\date{}

\begin{document}

\maketitle

\begin{abstract}
Small language models (SLMs) are attractive for agentic deployment due to low latency, reduced cost, and on-device privacy, yet they struggle with tool-use tasks where training data is scarce and noisy. Unlike larger models, SLMs cannot compensate for low-quality supervision through sheer capacity, making data quality the critical bottleneck. We present Data Turnstile, an open-source framework that takes user-defined API specifications and generates high-quality synthetic training data for function calling. Turnstile decomposes multi-turn tool-use interactions into constrained, stepwise generation with validation and error-feedback loops, providing fine-grained control over API diversity, conversation complexity, and output correctness. 
We demonstrate effectiveness of domain adaptation with Turnstile data on two challenging function calling benchmarks. 
On the BFCL single-turn benchmark, a Qwen3-0.6B fine-tuned on Turnstile data without chain-of-thought achieves 75.9\% overall accuracy (versus 67.4\% for the base model with thinking enabled), closing the gap with thinking-enabled Qwen3-1.7B (78.4\%) and Qwen3-4B (79.9\%) despite being 3$\times$ and 7$\times$ smaller respectively. 
On $\tau^2$-bench, a multi-turn agentic benchmark, Turnstile-trained Qwen3-1.7B achieves 31.1\% pass\textasciicircum{}1 on the Telecom domain, improving 4.7$\times$ over its 6.6\% base and surpassing Qwen2.5-32B-Instruct (27.4\%), a model 19$\times$ larger. 
Turnstile-trained Qwen3-0.6B achieves 24.6\%, improving 7$\times$ over its 3.5\% base and approaching the 32B model (53$\times$ larger). 
We release Data Turnstile along with a dataset spanning 1,000+ APIs and 100K+ multi-turn interactions.
\end{abstract}

\section{Introduction}
\label{sec:intro}

Function-calling\footnote{We use \emph{tool}, \emph{function}, and \emph{API} interchangeably.} language models power applications from customer service to coding assistants, but deploying them reliably requires expensive frontier models~\citep{fc_survey}.
Such models are typically large proprietary LLMs (32B+, often 100B+) and therefore deployed on the cloud.
In comparison, small language models (SLMs, $<$4B parameters) offer practical benefits for edge deployment: lower inference cost, reduced latency, and improved data privacy~\citep{lu-etal-2025-demystifying}.
However, these models face specific challenges in terms of their agentic ability: they struggle to (a) natively reason about and execute complex agentic workflows, and (b) generalize from noisy or imperfect training data that larger models can tolerate. Hence, improving function calling in SLMs is an active area of research~\citep{Belck2025SmallLM}.

LLMs typically excel at tool calling in a zero-shot setting, or with in-context learning through a few demonstrations~\citep{schick2023toolformer, patil2024gorilla}.
For SLMs however, supervised fine-tuning (SFT) on diverse, high-quality data is the primary path to competent agentic behavior~\citep{lin2024hammer, zhang-etal-2025-xlam}.
They must internalize conversational workflows, tool-calling patterns, API schemas, and multi-step reasoning directly from the training data.
Defects in training data are amplified disproportionately, as they cannot generalize from noisy data the way larger models can~\citep{Sorscher2022BeyondNS}. 
This makes data quality the binding constraint for training SLMs~\citep{Gunasekar2023TextbooksAA, liu2024apigen} and data generation methodology a first-class research problem.

\begin{figure}[t]
\centering
\includegraphics[width=0.55\textwidth]{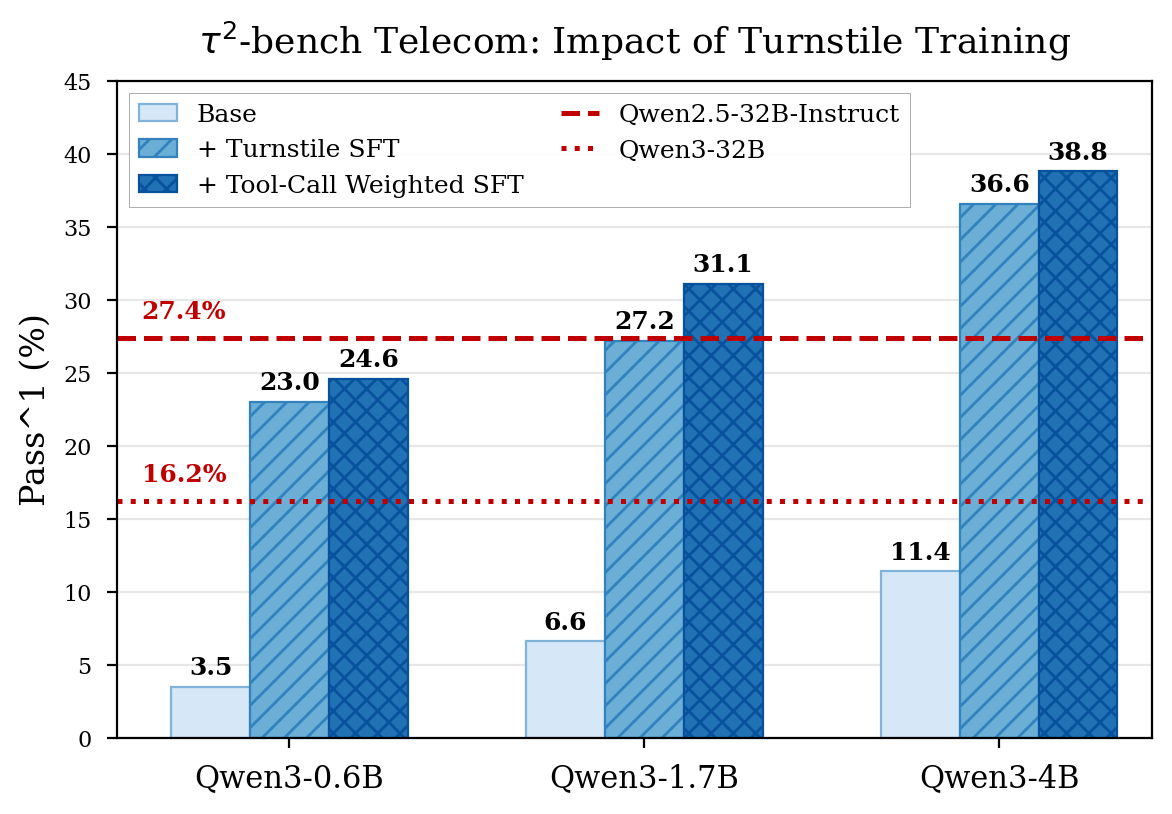}
\caption{
$\tau^2$-bench multi-turn results: Domain adaptation with Turnstile data helps SLMs beat zero-shot 32B models. 
}
\label{fig:hero}
\end{figure}

\paragraphbf{Gaps.} Existing approaches for generating synthetic tool-use data face several limitations.
End-to-end generation methods such as ToolBench~\citep{xu2023toolbench} typically prompt an LLM to produce entire multi-turn conversations in a `single-shot', i.e. through a single call to the LLM.
This leads to quality degradation in longer and complex interactions, with unnatural conversations, malformed API calls, hallucinated parameter values and illogical API execution outputs.
Execution-verified approaches such as APIGen~\citep{liu2024apigen} achieve higher quality but require functional API implementations, limiting to APIs with execution backends.
Most approaches also use proprietary frontier models for generation (e.g. ToolBench uses ChatGPT), making iteration and reproducibility difficult and expensive.

\paragraphbf{Our Approach.} We present \emph{Data Turnstile}, a  framework for generating high-quality synthetic training data for tool-use.
In contrast to single-shot generation, Turnstile decomposes each interaction into a directed acyclic graph (DAG) of typed `roles': user queries, reasoning traces, API calls, execution outputs, and assistant responses.
Each role is generated independently with focused constraints and validation, retrying with error feedback on failure.
This decomposition has two key benefits.
First, it imposes quality checks at each step to catch cascading errors.
Second, it reduces the per-step generation task to a complexity that cheap, open-weight models handle reliably, eliminating the need for proprietary LLMs.
This enables local execution and rapid iteration cycles at a fraction of the cost. 
Our methodology enables generation of training data of varying complexity, from simple single-turn user requests to nuanced, multi-step agentic workflows, while maintaining structural validation throughout.
\emph{We open-source the framework, enabling any team to generate targeted tool-use data for their own APIs.}

We demonstrate effectiveness on two challenging benchmarks.
On the BFCL single-turn function calling benchmark~\citep{patil2025bfcl}, a Qwen3-0.6B model trained on Turnstile data achieves 75.9\% accuracy (base: 67.4\%), within touching distance of the base Qwen3-4B-Instruct (79.9\%), a model almost 7$\times$ larger. 
On $\tau^2$-bench's multi-turn Telecom benchmark~\citep{barres2025tau2}, the Qwen3 1.7B and 4B trained on our data match or beat Qwen2.5-32B-Instruct (Figure~\ref{fig:hero}).
Additionally, Turnstile-trained SLMs achieve good performance even without chain-of-thought (CoT)~\citep{CoT}; avoiding the inference-time reasoning overhead for latency-sensitive deployments.
\paragraphbf{Contributions.} 
Our contributions are as follows:
\begin{itemize}
  \item We present Data Turnstile\footnote{\url{https://github.com/amazon-science/data-turnstile}}, a scalable open-source framework for generating function-calling data from custom APIs, using a step-wise method controlling quality and diversity. 
  \item We empirically show the efficacy of our approach, evaluating on BFCL and $\tau^2$-bench across three model scales (0.6B - 4B) with ablations on CoT and tool-call weighted SFT loss.
  \item We release a Turnstile-generated synthetic dataset~\footnote{\url{https://huggingface.co/datasets/amazon/Turnstile-Synthetic-Domains}} spanning 1K+ APIs and 100K+ multi-turn interactions with CoT reasoning traces, suitable for SLM fine-tuning.
\end{itemize}

\begin{figure*}[t]
  \centering
  \includegraphics[width=\textwidth]{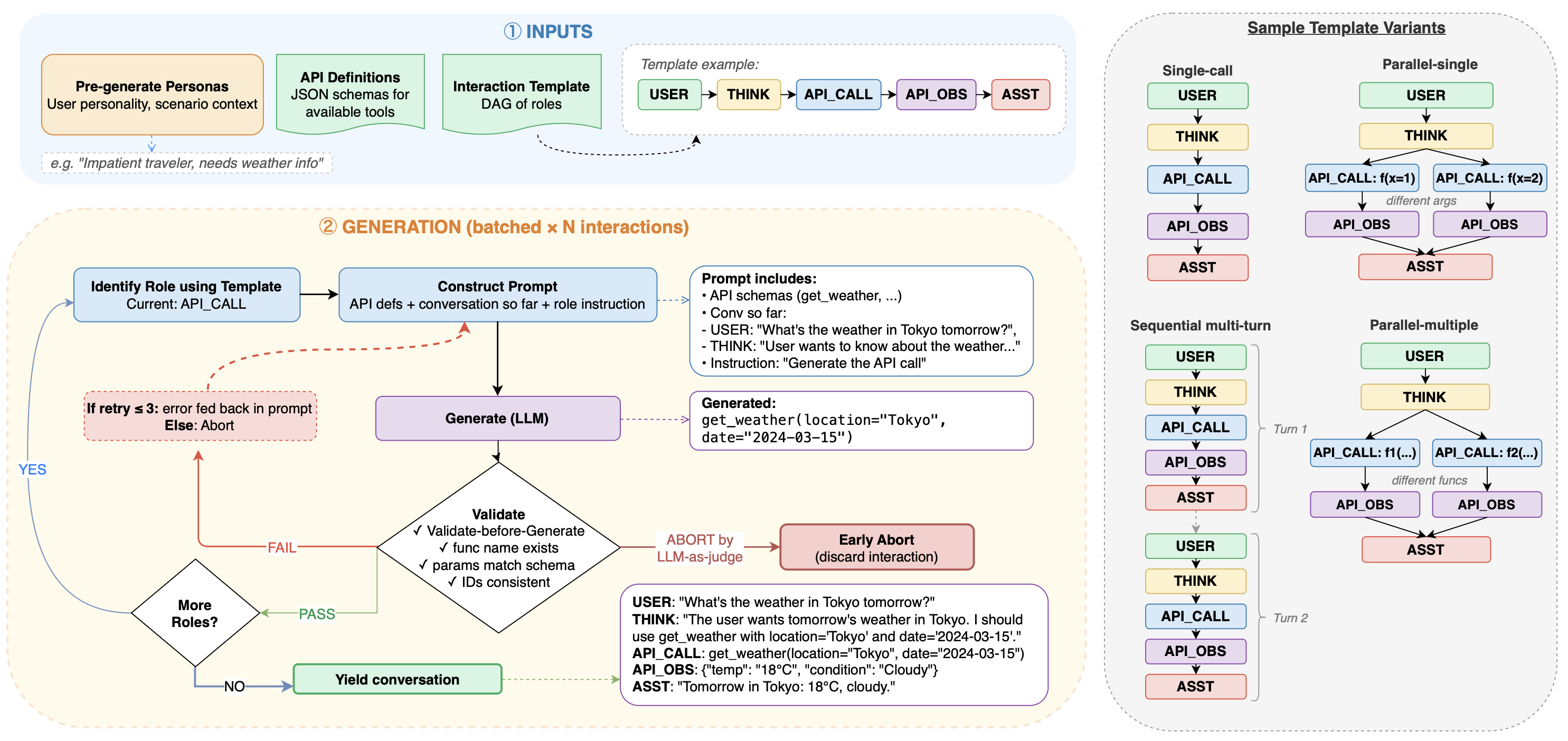}
  \caption{Data Turnstile Overview: role-wise generation with per-step validation and error-feedback based retry.}
  \label{fig:framework}
\end{figure*}

\section{Data Turnstile}
\label{sec:framework}
Generating high quality function-calling data is challenging: user requests must be realistic, API calls must be syntactically well-formed, API parameters must be grounded (hallucination-free), API execution outputs must be detailed and plausible, etc. 
The complexity compounds further when there are multiple user turns and API calls involved, as in real-world workflows. 
LLMs are well-known to solve complex tasks by decomposing them into tractable subproblems, whether implicitly through internal CoT reasoning, or explicitly through targeted decomposition~\citep{zhou2023leasttomost}.
We took inspiration from the latter approach to build Data Turnstile; by breaking down the problem of generating tool-use interactions into a set of independently constrained and validated generation steps.

\subsection{Template-Based Generation}
\label{sec:framework:templates}

The core abstraction is an \emph{interaction template}: a DAG $\mathcal{T} = (V, E, \Theta)$ where each node $v \in V$ corresponds to a \emph{role}, an atomic generation step. 
Directed edges $E$ encode dependencies between roles, while $\Theta$ defines additional generation specifications and context (API definitions, user personas, quality checks).
A template specifies the skeleton of an interaction by codifying its \emph{structure}, without the content. 
The interaction content itself varies with different parameter draws from $\Theta$, meaning a single template can yield many structurally identical but content-diverse interactions.

We define five roles for function-calling data: \texttt{USER}, \texttt{API\_CALL}, \texttt{API\_OBS} (tool responses), \texttt{ASSISTANT}, and \texttt{THINKING} (CoT reasoning). 
Generation proceeds sequentially through the template, as shown in Figure~\ref{fig:framework}.
At each step, we construct a prompt comprising: (1)~the structural context of the full template, (2)~the outputs of previously generated roles, and (3)~generation context specific to the current role. 
This means the language model has visibility into both what has been generated \emph{and} the outline of what is expected next. 
The DAG edges explicitly codify inter-role dependencies, ensuring each role receives exactly the context it needs while allowing the LLM to plan ahead. 
For instance, knowing that a specific \texttt{API\_CALL} will follow, it can craft a \texttt{USER} request which naturally elicits it.
After generation, the output is validated against role-specific constraints.
If successful, generation proceeds to the next role, otherwise it is retried with error feedback or aborted altogether as required. 
Multiple template instantiations are batched together for efficiency, with the generation state maintained independently for each. 

Quality and diversity are two key concerns for synthetic data. 
Interaction templates provide the ideal abstraction for \textit{structured diversity} with fine-grained quality controls, which we discuss next.

\subsection{Data Quality}
\label{sec:framework:quality}

\paragraphbf{Structural validations.}
The interaction template deterministically enforces format compliance at two levels: (a) sequence-level ordering (\texttt{API\_OBS} must always follow an \texttt{API\_CALL}; \texttt{THINKING} must precede \texttt{API\_CALL} or \texttt{ASSISTANT}) and (b) per-role validation (\texttt{API\_CALL} must conform to its schema definitions; every \texttt{API\_OBS} must be a well-formed JSON with the required fields).

\paragraphbf{Validate-before-Generate}
Before generating each role, the LLM validates the previous role content against predefined error criteria, catching issues beyond structural validation, such as hallucinated parameters and implausible API observations. 

\paragraphbf{Error Feedback and Early Abort.}
When structural or pre-generation validations fail for a role, it is retried up to a predefined budget, with additional context through an error feedback loop. 
This allows the LLM to rectify previous flaws, and potentially salvage the generation cost of the interaction thus far (instead of immediately discarding). 
If issues are irrecoverable, the LLM has the option to trigger an \textit{early abort}.
Together, these mechanisms ensure only high-quality interactions reach the final dataset.

\subsection{Data Diversity}
\label{sec:framework:diversity}
\paragraphbf{Template-level.}
Templates allow interactions with varying structural patterns (turns, number of API calls, etc.), with the distribution controlled with weights. For instance, we may want 50\% single-API single-turn, 25\% multi-API single-turn and 25\% multi-turn data. 

\paragraphbf{Parameter-level.}
Each template instance is associated with a particular $\Theta$, the set of generation specifications. 
These typically include the API names for every \texttt{API\_CALL}, a user persona to guide the request, and other information to inform the generation (like \textit{scenarios} discussed in \S~\ref{sec:taubench:generation}).
The combinatorial product of templates and parameters ensures \textit{structured diversity}, which cannot be achieved solely with temperature sampling.

\paragraphbf{Dynamic Template Perturbations.}
We introduce dynamic perturbations to the interaction template for an additional degree of data diversity. 
For example, tool execution failures can be introduced to elicit retry behavior from the model. 
We can simulate the incomplete information case, where the model is forced to ask for additional clarifications from the user before executing the API call. 
These perturbations are critical for generating more realistic data to train robust models that can recover from unexpected scenarios.

\subsection{Implementation}
\label{sec:framework:system}
Turnstile is implemented as a modular Python package with four core components: (a) \emph{Role}: Atomic units of interactions, encapsulating generation constraints and validations; (b) \emph{Template}: DAG specifications composed of multiple \emph{Roles} and their dependencies; (c) \emph{Builder}: Stateful controllers that orchestrate the generation loop (Figure~\ref{fig:framework}); (d) \emph{API Library}: Pluggable wrappers over a set of API definitions.
The specifications of a generation run are controlled by a top-level config, including template distribution, API library, teacher LLM, and other hyperparameters. 
The LLM itself is served by vLLM~\citep{kwon2023efficient}.
We built the initial framework from scratch with no AI assistance. 
For the open-source package, we used Qwen2.5-32B-Instruct to assist with code clean-up and documentation. 

\paragraphbf{Generation Efficiency.}
Per-role generation produces approximately the same number of total output tokens as single-shot generation. 
The overhead is additional LLM calls (one per role rather than one per interaction) and prompt prefills.
In practice, this overhead is modest: 
(1)~Each prefill is smaller than single-shot generation prompts, as only context specific to the current generation step is provided.
(2)~vLLM's paged attention~\citep{kwon2023efficient} ensures efficient KV-cache re-use across requests and steps. 
(3)~Continuous batching~\citep{continuous_batching} and chunked prefill~\citep{chunked_prefill} amortizes the prefill cost across decoding steps. 
(4)~The error retry mechanism drastically improves generation success (discussed in \S\ref{sec:bfcl:data_gen_analysis}).

\section{BFCL: Single Turn Function-Calling}
\label{sec:bfcl}

The Berkeley Function Calling Leaderboard (BFCL)~\citep{patil2025bfcl} is a popular benchmark for agentic models. We focus on the single-turn test sets, which contain six categories: \emph{Simple} (single tool, single API call expected), \emph{Multiple} (many tools, single API call expected), \emph{Parallel} (single tool, many API calls expected), \emph{Parallel Multiple} (many tools, many API calls expected), \emph{Relevance} (tools available but unnecessary), and \emph{Irrelevance} (no applicable tool).
For each test interaction, tool definitions are provided in the prompt, and correctness is evaluated via AST parsing of function calls.

\subsection{Single Turn Data Generation with Turnstile}
\label{sec:bfcl:generation}
In this section, we describe how Turnstile (\S~\ref{sec:framework}) is applied to generate single-turn function-calling data at scale, targeting general-purpose tool-use ability across arbitrary API definitions.

\paragraphbf{Templates.}
We define the following templates that closely mimic the BFCL categories: single-call (one API invoked), parallel-single (same API called $N$ times with different arguments), parallel-multiple ($N$ different APIs invoked simultaneously), and multi-turn (one API per turn over multiple turns).
Specific API names are assigned to template instances, by sampling from the global set of API definitions. 
In each turn, reasoning traces are generated following the user request to assess the impact of CoT on function-calling performance.
We implement a perturbation on the single-call template for generating irrelevance data, by generating the user request pertinent to a tool that is not available to the model. 

\paragraphbf{APIs and Datasets.}
We generate four large-scale synthetic datasets from different API sets: xLAM~\citep{zhang-etal-2025-xlam} and Glaive~\citep{glaive-function-calling-v2} (definitions extracted from open-source datasets), Synthetic Domains 
(synthetically generated APIs, details in \S\ref{sec:bfcl:synth_release}), 
and BFCL (definitions derived from the BFCL single-turn test set). 
Each source contains 1-3K definitions, providing a large pool for generation at scale.
The dataset statistics are summarized in Table~\ref{tab:dataset_stats}. We used Qwen2.5-32B-Instruct model as the generation LLM on a local p5.48xlarge instance (8 H100 GPUs). 
Turnstile's diversity levers produce datasets that are larger and measurably more diverse than raw open-source data from the same APIs (\S\ref{sec:bfcl:data_gen_analysis}). On average, Turnstile datasets have more roles and API calls than the open-source versions. 
This is due to the multi-turn multi-API templates used for generation, along with additional CoT reasoning traces present in our data. 

\definecolor{colhl}{HTML}{E8F0FE}

\begin{table}[!t]
\centering
\caption{Key Dataset Statistics. OS refers to open-source datasets, without CoT reasoning. All others are Turnstile generated and contain CoT traces.}
\label{tab:dataset_stats}
\small
\begin{tabular}{l|>{\columncolor{colhl}}r>{\columncolor{colhl}}r|ccc}
\toprule
&  &  & \multicolumn{3}{c}{\textbf{Per Interaction: avg {\scriptsize(min-max)}}} \\
\textbf{Domain} & \cellcolor{colhl}\textbf{Size} & \cellcolor{colhl}\textbf{\# APIs} & Roles & API Calls & Tokens \\
\midrule
xLAM {\scriptsize(OS)}           & $\sim$59K & 3,602 & 2.7 {\scriptsize(2-10)}   & 1.6 {\scriptsize(1-9)}   & 240 {\scriptsize(74-1.8K)}    \\
Glaive {\scriptsize(OS)}         & $\sim$78K & 1,069 & 5.9 {\scriptsize(2-14)}  & 1.7 {\scriptsize(1-2)}   & 279 {\scriptsize(76-2.1K)}   \\
\midrule
xLAM           & $\sim$230K & 3,178 & 7.0 {\scriptsize(5-10)}   & 1.7 {\scriptsize(1-2)}   & 520 {\scriptsize(73-2.5K)}    \\
Glaive         & $\sim$199K & 1,064 & 11.2 {\scriptsize(6-16)}  & 2.7 {\scriptsize(1-4)}   & 623 {\scriptsize(109-1.8K)}   \\
Synth. & $\sim$100K & 1,025 & 9.9 {\scriptsize(5-15)}  & 2.8 {\scriptsize(1-4)}  & 972 {\scriptsize(235-6.8K)}   \\
BFCL           & $\sim$159K & 1,260 & 9.6 {\scriptsize(6-26)}   & 2.0 {\scriptsize(1-5)}   & 540 {\scriptsize(110-2.9K)}   \\
Telecom        &  $\sim$36K &    14 & 38.9 {\scriptsize(17-81)} & 4.1 {\scriptsize(2-13)}  & 1.6K {\scriptsize(649-3.1K)} \\
\bottomrule
\end{tabular}
\end{table}

\paragraphbf{Distractors.}
During training, we augment each interaction with semantically similar distractor APIs alongside the ones present in the interaction, forcing the model to discriminate between plausible alternatives and select the appropriate tool.

\subsection{Experiment Setup}
\label{sec:bfcl:exp}

\paragraphbf{Models.}
We fine-tune Qwen3-0.6B on different combinations of function-calling data, comparing against Qwen3 base models~\citep{yang2025qwen3technicalreport} at three scales (0.6B, 1.7B, 4B) without fine-tuning. 
We evaluate the models in two inference modes: \emph{think} (explicit CoT before function calls) and \emph{no-think} (direct output), to assess the performance gap.

\definecolor{colhl}{HTML}{E8F0FE}
\definecolor{sechl}{HTML}{F2F2F2}

\begin{table*}[!ht]
\centering
\caption{BFCL single-turn results on Qwen3 models. \textit{Raw OS}: unprocessed open-source xLAM+Glaive data. \textit{Turnstile-OS}: Turnstile-generated data from xLAM+Glaive APIs. \textit{Turnstile-OOD:} Turnstile-OS + data from Synthetic Domain APIs. \textit{Turnstile (OOD+ID)}: OOD + adds data generated from BFCL's own API definitions. \textbf{Bold} = best overall; \underline{underline} = best among SFT models.}
\label{tab:single-turn}
\resizebox{\textwidth}{!}{%
\begin{tabular}{l|>{\columncolor{colhl}}c>{\columncolor{colhl}}c>{\columncolor{colhl}}c|cc|cccc|cccc}
\toprule
& \multicolumn{3}{c|}{\cellcolor{colhl}\textbf{Overall}} & \multicolumn{2}{c|}{\textbf{Detection}} & \multicolumn{4}{c|}{\textbf{Non-Live (AST)}} & \multicolumn{4}{c}{\textbf{Live (AST)}} \\
\textbf{Model} & \cellcolor{colhl}\textbf{Avg} & \cellcolor{colhl}\textbf{Non-Live} & \cellcolor{colhl}\textbf{Live} & \textbf{Rel.} & \textbf{Irrel.} & Simple & Multi & Para & Para-M & Simple & Multi & Para & Para-M \\
\midrule
\rowcolor{sechl}
\multicolumn{14}{l}{\textit{Base Models (think eval)}} \\
\midrule
Qwen3-4B & \textbf{79.9} & \textbf{87.1} & 72.6 & 88.2 & 75.3 & \textbf{78.9} & 93.5 & \textbf{89.5} & \textbf{89.5} & 80.2 & \textbf{76.1} & 62.5 & 66.7 \\
Qwen3-1.7B & 78.4 & 81.4 & \textbf{75.3} & 52.9 & 83.3 & 72.3 & 89.5 & 80.5 & 81.0 & 74.8 & 69.8 & 43.8 & \textbf{79.2} \\
Qwen3-0.6B & 67.4 & 72.3 & 62.5 & 52.9 & 81.0 & 63.3 & 75.0 & 74.0 & 66.5 & 60.9 & 49.3 & 62.5 & 45.8 \\
\midrule
\rowcolor{sechl}
\multicolumn{14}{l}{\textit{Base Models (no-think eval)}} \\
\midrule
Qwen3-4B & 71.1 & 77.6 & 64.5 & 82.4 & 52.6 & 74.9 & 90.0 & 87.5 & 84.0 & \textbf{81.4} & 69.2 & \textbf{68.8} & 66.7 \\
Qwen3-1.7B & 67.0 & 76.3 & 57.7 & \textbf{100.0} & 56.2 & 72.9 & 90.5 & 80.5 & 76.0 & 70.9 & 59.8 & 43.8 & 58.3 \\
Qwen3-0.6B & 58.2 & 63.0 & 53.4 & 76.5 & \textbf{90.6} & 57.0 & 62.5 & 54.5 & 49.5 & 38.4 & 27.5 & 12.5 & 29.2 \\
\midrule
\rowcolor{sechl}
\multicolumn{14}{l}{\textit{SFT-0.6B-no-think: Methodology ablation (same xLAM+Glaive APIs)}} \\
\midrule
Raw-OS & 55.1 & 63.9 & 46.2 & 76.5 & 35.7 & 67.2 & 84.5 & 73.0 & 65.5 & 60.5 & 46.3 & 31.3 & 45.8 \\
Turnstile-OS & 70.4 & 75.9 & 65.0 & 88.2 & 77.5 & 70.9 & 89.0 & 75.0 & 62.0 & 59.9 & 62.5 & \underline{62.5} & 25.0 \\
\midrule
\rowcolor{sechl}
\multicolumn{14}{l}{\textit{SFT-0.6B-no-think: Different API Sources}} \\
\midrule
Turnstile-OOD & 72.9 & 78.6 & 67.2 & \underline{76.5} & 80.2 & 72.0 & 92.0 & 75.5 & 70.5 & 60.5 & 61.5 & 50.0 & 16.7 \\
Turnstile-OOD+ID & \underline{75.9} & \underline{81.5} & \underline{70.5} & 70.6 & \underline{81.8} & \underline{72.9} & \textbf{\underline{94.0}} & 76.5 & \underline{78.0} & \underline{68.6} & \underline{65.9} & 43.8 & \underline{45.8} \\
\midrule
\rowcolor{sechl}
\multicolumn{14}{l}{\textit{SFT-0.6B-think (trained in thinking mode; think eval)}} \\
\midrule
Turnstile-OOD & 69.7 & 76.4 & 63 & \underline{76.5} & 77.0 & 69.9 & 91.0 & 76.0 & 65.0 & 58.1 & 56.1 & 43.8 & 16.7 \\
Turnstile-OOD+ID & 72.8 & 79.8 & 65.8 & 70.6 & \underline{81.8} & 70.8 & \textbf{\underline{94.0}} & \underline{79.5} & 75.0 & 66.7 & 62.7 & 43.8 & 29.2 \\
\bottomrule
\end{tabular}%
}
\end{table*}

\paragraphbf{Experiments.}
We design our experiments to investigate single-turn performance, and evaluate the following: 
\begin{enumerate}
    \item Impact of training on existing open-source datasets (xLAM-OS and Glaive-OS). This is a baseline to compare Turnstile SFT against.
    \item Impact of out-of-distribution (OOD) Turnstile data, generated using xLAM, Glaive, and Synthetic Domains APIs\footnote{OOD APIs have just ~2\% overlap with BFCL evaluation APIs.}, measuring whether high-quality tool calling data improves performance on unseen evaluation APIs.
    \item Impact of in-distribution (ID) Turnstile data, generated from BFCL's APIs, to assess whether training on the target API schemas provides additional benefit. This is particularly relevant for SLMs deployed on a fixed set of domain APIs.
\end{enumerate}

\paragraphbf{Training Details.}
All models were trained identically, with datasets sampled proportional to their volume. 
We perform full fine-tuning using AdamW with a learning rate of $5{\times}10^{-5}$, linear warmup over 50 steps, an effective batch size of 64, and a maximum sequence length of 4096.
We train in both no-think and think modes; each model is evaluated in its corresponding training mode.

% -----------------------------------------------------------------------------
\subsection{Results}
\label{sec:bfcl:results}
Table~\ref{tab:single-turn} presents our main single-turn results. BFCL contains two splits: `Non-live' (synthetic APIs) and `Live' (real-world APIs).

\paragraphbf{Turnstile SFT vs.\ base models}
(Rows: SFT no-think vs.\ Base think/no-think.)
Turnstile (OOD+ID) no-think achieves 75.9\%, surpassing the base 0.6B in both modes: think (67.4\%) and no-think (58.2\%). This is notable for SLM deployment since Turnstile SFT without CoT exceeds performance that the base model can only achieve with CoT overhead. It also approaches the 3$\times$ larger Qwen3-1.7B (78.4\%) and the 7$\times$ larger Qwen3-4B (79.9\%) base models with thinking.

\paragraphbf{Impact of generation methodology} (Rows: Turnstile-OS vs Raw-OS) The most controlled comparison is between Raw-OS and Turnstile-OS, since both draw from the same xLAM and Glaive API schemas; only the generation methodology differs. Turnstile-OS achieves 70.4\% versus 55.1\% for Raw-OS, a gain of +15.3 percentage points (pp) attributable to generation methodology alone. The base 0.6B already scores 67.4\% with thinking; open-source SFT \emph{degrades} this native ability to 55.1\%, whereas Turnstile lifts it well beyond without CoT overhead. 
Per-category analysis reveals two failure modes of open-source data: irrelevance detection collapses from 81\% (in base 0.6B) to 35.7\% (every training sample invokes a tool, so the model never learns to refuse), and live accuracy drops to 46.2\%. Turnstile addresses both through dynamic irrelevance injection (recovering refusal to 77.5\%) and template-controlled structural diversity (65.0\% on live APIs without training on those schemas).

\paragraphbf{Impact of API diversity and target alignment}
(Rows: SFT no-think, Turnstile-OOD and OOD+ID.)
To study the effect of API diversity, we used the Synthetic Domains dataset generated with Turnstile.
Adding this to Turnstile-OS (forming Turnstile-OOD) yields a +2.5pp gain over Turnstile-OS (70.4\%$\to$72.9\%), with improvements across non-live and live splits. The broader API coverage helps the model generalize to unseen evaluation schemas. Further adding Turnstile data generated from BFCL's own API schemas (Turnstile-OOD+ID) yields +3pp (72.9\%$\to$75.9\%), distributed across non-live (+2.9\%) and live (+3.3\%), with the largest gain on parallel-multiple (70.5\%$\to$78.0\%). OOD data teaches the core function-calling skill across diverse APIs, while ID data provides incremental alignment to the target schema conventions. 
This validates Turnstile's practical utility for deployment, where practitioners can generate targeted data for their own domain and use cases.

\paragraphbf{Impact of CoT}
(Rows: SFT think vs.\ SFT no-think, Turnstile OOD+ID.)
No-think Turnstile (OOD+ID) outperforms its think counterpart (75.9\% vs.\ 72.8\%), with the gap largest on live APIs (70.5\% vs.\ 65.8\%) and parallel-multiple patterns (45.8\% vs.\ 29.2\% live). Base models show the opposite pattern, with all three dropping significantly without CoT (0.6B: 67.4\%$\to$58.2\%, 1.7B: 78.4\%$\to$67.0\%, 4B: 79.9\%$\to$71.1\%), indicating a learned dependence on reasoning for tool-calling. Turnstile SFT removes this dependency entirely. We identify two failure modes of CoT in single-turn settings: \emph{irrelevance rationalization} (reasoning into calling inapplicable APIs) and \emph{parameter overthinking} (over-interpreting ambiguous constraints on parallel calls). A qualitative analysis is provided in Appendix~\ref{sec:appendix:think}.

\subsection{Released Dataset: Synthetic Domains}
\label{sec:bfcl:synth_release}

We publicly release the Synthetic Domains dataset under the CC BY-NC 4.0 license.
It comprises ${\sim}$100K Turnstile-generated interactions over 1,025 APIs spanning 50+ categories (finance, weather, maps, e-commerce, etc.). 
It covers single-call, parallel, parallel-multiple, and multi-turn templates, with CoT reasoning traces.

\paragraphbf{Standalone evaluation.}
Training Qwen3-0.6B on this dataset alone achieves 67.4\% on BFCL single-turn, surpassing both the unprocessed open-source baseline (Raw-OS, 55.1\%) and the base 0.6B no-think model (58.2\%). This demonstrates that the released data is directly useful for improving SLM tool-calling ability.

\paragraphbf{Human assessment.}
Two annotators rated 100 random interactions on \emph{correctness}, \emph{naturalness}, and \emph{groundedness} using a Likert scale of 1--5. On average, we obtained scores of 4.4, 3.8, and 4.5 respectively, indicating overall high quality of the synthetic data.

\subsection{Data Generation: Analysis}
\label{sec:bfcl:data_gen_analysis}
In this section, we discuss insights we gathered during the generation of the Synthetic Domains dataset.

\paragraphbf{Impact of Error Retries.}
We started with a seed size of 125K template instantiations, to generate $\sim$100K interactions ($\sim$84\% success rate).
Of the successful interactions, $\sim$22\% encountered at least one error during the step-wise generation, but eventually succeeded due to Turnstile's \emph{retry with error feedback}.
Therefore this mechanism is critical for maintaining higher success rates, and enabling local generation with smaller less powerful models (like Qwen-32B). 
Interestingly, 87\% of the errors flagged were through structural validations (e.g. invalid API calls) while the rest were through the per-role LLM-as-Judge (primarily catching hallucinations). 

\definecolor{colhl}{HTML}{E8F0FE}

\begin{table}[t]
\centering
\caption{Diversity comparison between Turnstile-generated data and open-source data. All metrics are in range [0, 1]; 1.0 indicates maximum diversity. Tools $\geq$10 shows the percentage of tools qualifying for Arg Richness computation.
}
\label{tab:diversity}
\small
\begin{tabular}{l|>{\columncolor{colhl}}c>{\columncolor{colhl}}c>{\columncolor{colhl}}c>{\columncolor{colhl}}c|r}
\toprule
& \shortstack{\textbf{Tool}\\\textbf{Balance}} & \shortstack{\textbf{Call}\\\textbf{Seq.}} & \shortstack{\textbf{Structure}\\~} & \shortstack{\textbf{Arg}\\\textbf{Richness}} & \shortstack{\textbf{Tools}\\$\geq$10 (\%)} \\[5pt]
\midrule
xLAM \scriptsize{(OS)} & 0.92 & 0.91 & 0.85 & 0.84 & 71\% \\
Glaive \scriptsize{(OS)} & 0.63 & 0.67 & 0.62 & 0.61 & 26\% \\
\midrule
xLAM & \textbf{0.99} & \textbf{0.95} & 0.91 & 0.89 & \textbf{99\%} \\
Glaive & 0.98 & 0.94 & \textbf{0.96} & 0.89 & \textbf{100\%} \\
Synth. Dom. & \textbf{0.99} & 0.94 & 0.95 & \textbf{0.98} & \textbf{99\%} \\
\bottomrule
\end{tabular}
\end{table}

\paragraphbf{Failure rates vs complexity.}
Even with role-wise generation in Turnstile, we find that the failure rates increase drastically with increase in the complexity of the interaction. 
For single-turn interactions, the failure rate increases with the number of function calls: 7\%, 16\%, 18\% and 22\% for 1, 2, 3 and 4 parallel calls respectively. 
In multi-turn interactions (with one function call per turn), we see failure rates of 7\%, 9\% and 13\% for 1, 2 and 3 turns respectively.

\paragraphbf{Diversity analysis.}
We quantify the diversity of Turnstile data against the open-source datasets (xLAM, Glaive) using four metrics (Table~\ref{tab:diversity}, definitions in Appendix~\ref{sec:appendix:diversity}):
(a) \emph{Tool Balance}: the normalized entropy of API call frequencies, capturing how evenly tools are exercised; (b) \emph{Call Sequence}: the normalized entropy over the distribution of per-example tool-call orderings; (c)
\emph{Structure}: 4-gram normalized entropy over role sequences (USER, THINK, API\_CALL, API\_OBS, ASST), capturing variety in interaction patterns; (d) \emph{Arg Richness}: argument-value variety per tool. It is the fraction of unique argument-value combinations when sampling $n$ calls per tool (repeated 50 times), averaged across all qualifying tools. We choose $n{=}10$ to retain sufficient qualifying tools for comparison in OS data. 

Turnstile data consistently outperforms open-source data across all four metrics. Tool Balance metric indicates near-uniform API usage in Turnstile, while open-source (OS) shows skewed call distributions.
The gaps on Call Sequence (0.94-0.95 vs 0.67-0.91) and Structure (0.91-0.96 vs 0.62-0.85) highlight the limited structural patterns in OS data.
Arg Richness shows a similar trend, with Turnstile achieving 0.89-0.98 compared to 0.61-0.84 for OS data.
This validates that our template-based generation 
produces measurably more diverse data, even when drawing from the same API pools.

\begin{figure*}[t]
\centering
\includegraphics[width=\textwidth]{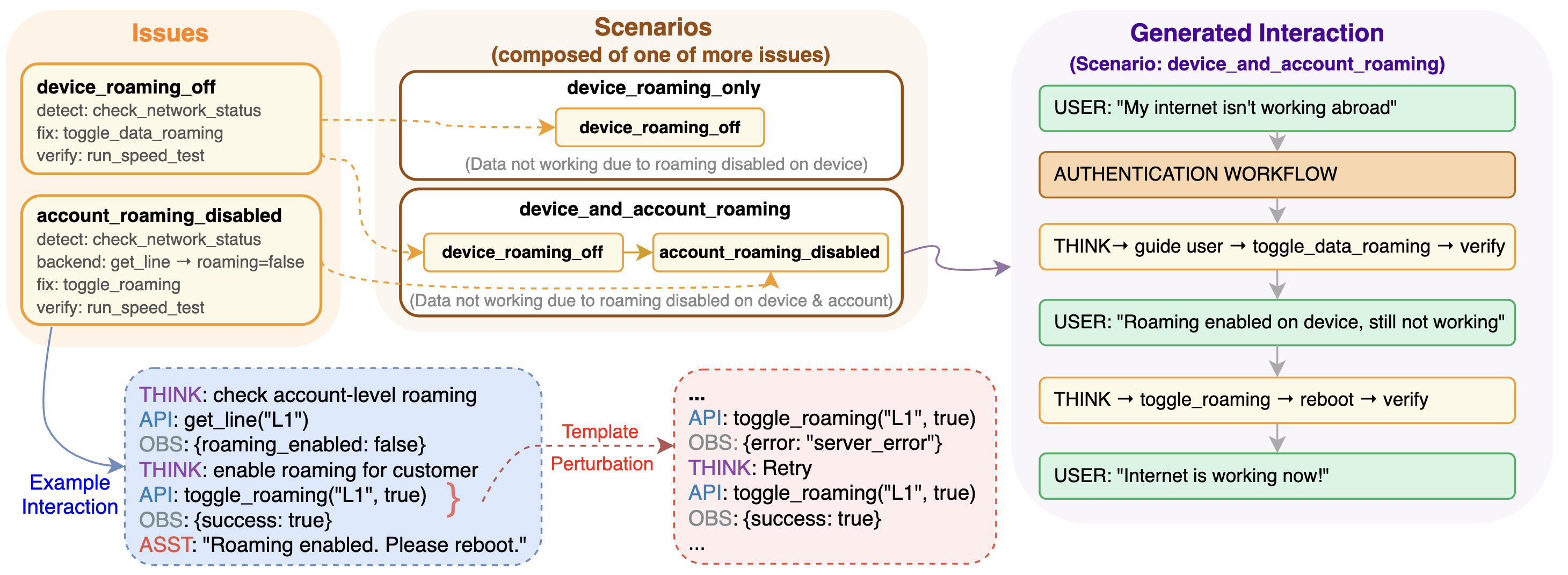}
\caption{Multi-turn data generation for $\tau^2$-bench Telecom: reusable \emph{Issues} compose into \emph{Scenarios} with dynamic perturbations.}
\label{fig:mt_framework}
\end{figure*}

\section{$\tau^2$-bench: Multi Turn Agentic Workflows}
\label{sec:taubench}

$\tau^2$-bench~\citep{barres2025tau2} is a multi-turn conversational benchmark that evaluates agentic tool use with user turns simulated by LLMs. 
It tests policy-adherent tool use, multi-step reasoning, and error recovery in realistic customer-service scenarios\footnote{We chose $\tau^2$-bench over the BFCL multi-turn benchmark, to demonstrate the capabilities of Turnstile in the setting of adherence to a predefined policy document}.
It provides 4 evaluation domains: Banking, Retail, Airline and Telecom. 
We focus on the Telecom domain for our experiments, comprising 114 tasks requiring diagnosis and resolution of various issues. 
Evaluations report pass\textasciicircum{}1, the percent of successful tasks averaged across four trials. 

% -----------------------------------------------------------------------------
\subsection{Policy-Based Generation with Turnstile}
\label{sec:taubench:generation}
There are three key differences between the $\tau^2$-bench Telecom and single-turn BFCL benchmark: (1) agent is expected to adhere to a policy document, (2) only 14 APIs in comparison to 1K+, and (3) much longer interactions, with emphasis on conversational ability for facilitating user-side debugging along with agent-side tools.

\paragraphbf{Issue and Scenario}
We define two new concepts to codify and enable policy-driven data generation. 
An \emph{Issue} minimally specifies a user problem, diagnosis steps/tools, and actions to fix it, typically consisting of 4-8 generation \emph{Roles}.
For example, the issue of 'Data Roaming Disabled' encapsulates the relevant root cause, diagnosis steps and fixes.
A \emph{Scenario} is a curated composition of one or more issues, with shared workflows (such as authentication) and additional generation context. 
Figure~\ref{fig:mt_framework} shows two separate scenarios with the same problem symptom (\emph{`Data not working'}), but with one (\emph{`device\_roaming\_only'}) or two (\emph{`device\_roaming\_only'} + \emph{`device\_and\_account\_roaming'}) issues as actual root causes.
In other words, \emph{Issues} are building blocks that can be used to compose scenarios that simulate real-world complexity. 
We derive a total of 17 issues and 34 composed scenarios from the policy document to construct a diverse and representative set of workflows. 

\paragraphbf{Quality and Diversity}
Multi-turn templates are dynamically assembled from the scenario and its constituent issues. 
We generate detailed reasoning traces to help the model understand the required assistant workflows. 
For additional diversity, we use template perturbations for simulating API execution failures, asking for missing information, and user personas (un-cooperative, tech-illiterate, etc.). 
Figure~\ref{fig:mt_framework} shows an example of the API execution failure, as applied to an interaction. 
Additionally, we chain multiple scenarios together to create more complex interactions, e.g. by joining with an \emph{`Is there anything else I can help with?'} assistant turn.

\paragraphbf{Dataset.}
We generated training data for the Telecom domain spanning the 14 APIs. 
The dataset comprises $\sim$35K interactions with an average of 39 roles per conversation and an average sequence length of 1.6K tokens (Table~\ref{tab:dataset_stats}). 
Note that this is substantially longer ($3$--$4\times$) and more complex than the single-turn data, as the model needs to maintain state and reason over extended context.

\begin{table}[pt!]
\centering
\caption{Multi-turn results on $\tau^2$-bench Telecom (114 tasks, 4 trials, pass\textasciicircum{}1 {\small [95\% CI]} ). SFT models trained on Turnstile Telecom data. Weighted SFT applies 5$\times$ loss on tool-call tokens. No-think models are trained/evaluated without CoT.}
\label{tab:multi-turn-results}
  \begin{tabular}{lcccc}
  \toprule
  \textbf{Model} & \textbf{Base} & \textbf{\cellcolor{colhl}+ SFT} & \textbf{\cellcolor{colhl}\shortstack{+ Tool-call \\ Weighted}} & \textbf{\shortstack{No-think \\
   SFT}} \\
  \midrule
  Qwen2.5-32B & 27.4 {\small [±6.5]} & \cellcolor{colhl}--- & \cellcolor{colhl}--- & --- \\
  Qwen3-32B & 16.2 {\small [±5.7]} & \cellcolor{colhl}--- & \cellcolor{colhl}--- & --- \\
  \midrule
  Qwen3-4B & 11.4 {\small [±3.9]} & \cellcolor{colhl}36.6 {\small [±6.1]} & {\cellcolor{colhl}\textbf{38.8} {\small [±6.5]}} & 14.5 {\small
  [±4.3]} \\
  Qwen3-1.7B & 6.6 {\small [±2.7]} & \cellcolor{colhl}27.2 {\small [±6.1]} & \cellcolor{colhl}31.1 {\small [±5.9]} & 16.0 {\small [±4.5]} \\
  Qwen3-0.6B & 3.5 {\small [±1.9]} & \cellcolor{colhl}23.0 {\small [±5.5]} & \cellcolor{colhl}24.6 {\small [±5.0]} & 11.8 {\small [±4.4]} \\
  \bottomrule
  \end{tabular}
\end{table}

\subsection{Experimental Setup}
\label{sec:taubench:setup}

We fine-tune Qwen3 models at three scales (0.6B, 1.7B, 4B) on Turnstile-generated Telecom data and evaluate against base models at the same scales, as well as Qwen3-32B and Qwen2.5-32B-Instruct which serve as upper-bound references without any fine-tuning.
We use Qwen2.5-32B-Instruct as the user simulator LLM, as no instruction-tuned Qwen3-32B variant is publicly available.

\paragraphbf{Training Details.}
We perform SFT using AdamW with a learning rate of $5{\times}10^{-5}$, batch size of 64, and a maximum sequence length of 8192. 
DeepSpeed ZeRO Stage 2~\citep{deepspeed} is used for 1.7B and 4B models. 
All models are trained with CoT/thinking enabled until convergence, determined via checkpoint sweeps on validation set. 
We also experiment with a \emph{tool-call weighted} variant that applies 5$\times$ loss weight (multiplier selected via validation) on the \emph{API call} tokens.
This is to help the model to prioritize correct API invocations, which are a small but crucial fraction of total tokens.

\subsection{Results}
\label{sec:taubench:results}

Table~\ref{tab:multi-turn-results} and Figure~\ref{fig:hero} summarize our multi-turn results.

\paragraphbf{Impact of Domain Adaptation.}
(Columns: Base vs. SFT).
All three models benefit from the SFT, with absolute gains of +19.5pp (0.6B), +20.6pp (1.7B), and +25.2pp (4B) over their respective baselines (with non-overlapping 95\% CIs). 
Relative improvements are larger for smaller models (6.5$\times$ for 0.6B vs.\ 3.2$\times$ for 4B), confirming the model capacity bottleneck. 
Notably, Qwen3-4B (36.6\%) surpasses Qwen2.5-32B-Instruct (27.4\%) by +9.2pp despite being 8$\times$ smaller, and Qwen3-1.7B (27.2\%) achieves parity at 19$\times$ smaller. 
Even the 0.6B model (23\%) outperforms Qwen3-32B (16.2\%) by +6.8pp. 
These gains come from 35K interactions generated entirely from a 14-API policy specification with no human-written examples, validating Turnstile's practical value for domain adaptation. 
One can specify their APIs and domain policy, generate data with Turnstile, and obtain a competent small model without manual data collection or annotation.

\paragraphbf{Tool-call Weighted SFT}
(Columns: SFT vs. Tool-call Weighted).
Tool-call weighted SFT shows consistent gains of +1.6 to +3.9pp across all three scales. While individual deltas fall within the 95\% CIs, the positive impact suggests this direction is worth exploring.

\paragraphbf{Impact of CoT}
(Columns: SFT vs No-think SFT)
In contrast to the single-turn finding, multi-turn agentic tasks benefit substantially from CoT reasoning. Without thinking, all models collapse by 11-22pp. Thinking enables three capabilities critical for multi-turn diagnosis: (1) \emph{hypothesis formation} (reasoning about possible causes before acting), (2) \emph{evidence-based progression} (interpreting API results to confirm or reject hypotheses), and (3) \emph{multi-issue discovery} (after resolving one problem, reasoning that additional issues may exist). Without this deliberative structure, models iterate blindly through available data, latch onto irrelevant signals, and terminate prematurely. A qualitative comparison is provided in Appendix~\ref{sec:appendix:think}.

\paragraphbf{Comparison to Single-Shot.}
We performed a controlled ablation to isolate the impact of our methodology, compared to single-shot generation.
We generated $\sim$35k interactions by mirroring the constraints provided to role-wise generation and enforcing validation checks post-hoc. 
We found major quality issues in the generated data: mainly stemming from oracle information leakage and false success declarations.
With SFT on this data, the SLMs show major degradations, with 1.7B and 4B models achieving just 3.5\% and 3.7\% respectively. 
More details are provided in Appendix~\ref{sec:appendix:single-shot}.

\begin{figure*}[pt!]
\centering
\includegraphics[width=\textwidth]{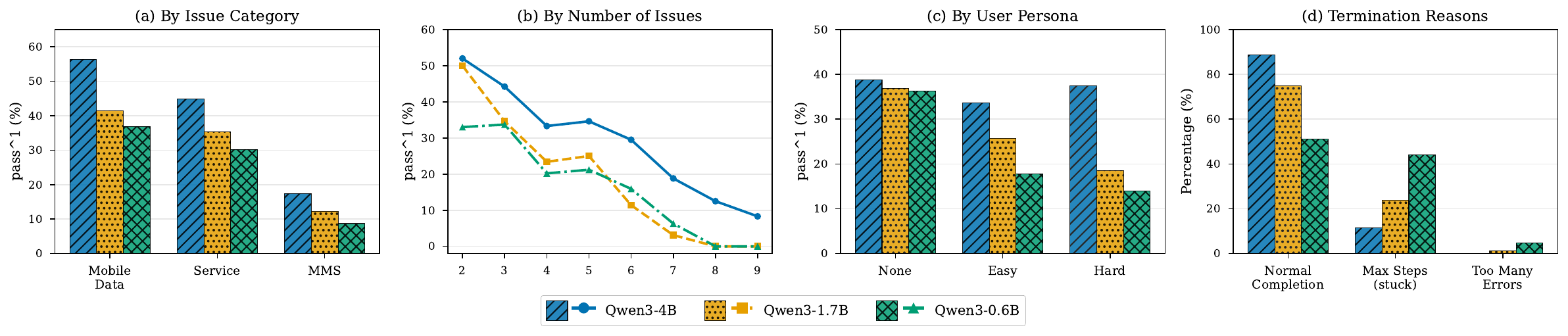}
\caption{$\tau^2$-bench Telecom: Error Analysis across different dimensions for Turnstile SFT models (0.6B, 1.7B, 4B).}
\label{fig:mt_Telecom_errors}
\end{figure*}

\subsection{Error Analysis: Discussion}
\label{sec:taubench:discussion}
We further analyze the model performance across model sizes on the 114 evaluation tasks, breaking it down by category (Figure~\ref{fig:mt_Telecom_errors}).

\paragraphbf{Category.}
Complaints span three categories (Mobile Data, Service, MMS) with a clear difficulty hierarchy: models resolve 40-60\% of Mobile Data tasks but perform poorly on MMS, which requires more complex debugging. 
We expect that richer multi-issue scenarios for harder categories would improve performance further. 
We could not invest further in this due to budget constraints.

\paragraphbf{Complexity.} Pass rate degrades monotonically with issue count per task, yet models generalize beyond training (which contains at most 3-issue scenarios) to solve tasks with up to 5 issues. The capacity gap widens with complexity: at 7 issues, 4B retains 18.8\% while 1.7B and 0.6B collapse to 3.1\% and 6.3\%. $\tau^2$-bench also evaluates across different user personas (None, Easy and Hard). They reveal a similar capacity threshold: 4B is robust to adversarial users (37.5\% Hard vs 38.8\% None), while 0.6B drops sharply from 36.3\% to 13.9\%. The models perform comparably with cooperative users.

\paragraphbf{Termination Behavior.}
Termination reasons reveal differences in failure modes across model sizes. The 4B model terminates normally in 88.6\% of cases, while 0.6B gets stuck in hallucination loops (44.1\%) or hits the error limit due to malformed tool calls (4.8\%). This suggests 0.6B lacks sufficient capacity for multi-turn reasoning over long context; 1.7B and 4B are relatively better suited.

\section{Related Work}
\label{sec:related_work}

\paragraphbf{Synthetic Data for Tool-Use.}
Training LLMs for function calling increasingly relies on synthetic data generated by stronger models~\citep{selfinstruct}.
ToolBench~\citep{xu2023toolbench} pioneered large-scale generation using depth-first decision tree reasoning across 16K real REST APIs with execution-based validation, while APIGen~\citep{liu2024apigen} demonstrated that post-hoc three-stage verification (format, execution, semantic) on 60K examples outperforms larger unfiltered corpora.
The field has since expanded to multi-turn settings: APIGen-MT~\citep{prabhakar2026apigenmt} uses blueprint-to-trajectory generation with LLM committee scoring, Magnet~\citep{yin-etal-2025-magnet} and ToolFlow~\citep{wang-etal-2025-toolflow} employ graph-based dependency modeling, and FunReason-MT~\citep{xu2025funreasonmttechnicalreportadvanced} introduces Environment-API Graph Interactions with guided chain-of-thought.
ToolWeave~\citep{khandelwal2026toolweavestructuredsynthesiscomplex} demonstrated that open-weight models can serve as generation agents for controllable multi-turn synthesis.
However, none of these works support generation from domain policy documents for deployment-specific API sets, and most rely on organic LLM variation for diversity rather than providing controllable guarantees.
Turnstile adopts execution-free validation with per-role error-feedback retry and explicit template-controlled diversity.

\paragraphbf{Open-Source Data Generation.}
Despite rapid progress, reproducibility remains limited. Only ToolBench releases usable pipeline code (requiring OpenAI API plus live API backends). FunReason-MT releases only model weights without its synthesis pipeline.
Among open-source tools, Distilabel~\citep{distilabel-argilla-2024} and the Azure Function-Calling Data Synthesizer~\citep{azure-fc-synthesizer-2025} reimplement APIGen-style pipelines but are restricted to single-turn generation.
CAMEL-AI~\citep{li2023camel} offers multi-agent data generation via role-playing but lacks per-turn structural validation.
To our knowledge, Turnstile is the first open-source framework that jointly supports multi-turn generation, per-role quality validation with retry, template-controlled diversity, and policy-adherent generation from domain specifications.

\paragraphbf{Benchmarks.}
We evaluate on BFCL v3~\citep{patil2025bfcl} for single-turn function calling and $\tau^2$-bench~\citep{barres2025tau2} for multi-turn agentic conversations with policy adherence.
Other benchmarks such as NESTFUL~\citep{basu-etal-2025-nestful} (nested compositional calls), T-Eval~\citep{chen-etal-2024-eval} (multi-step orchestration), and GTA~\citep{wang2024gta} (executable tool chains) highlight that multi-turn, policy-adherent tool use remains unsolved even for frontier models.

\section{Conclusion}
\label{sec:conclusion}

We presented Data Turnstile, a framework for generating high-quality synthetic tool-use data through step-wise role generation and validation. It supports both general function-calling (from custom API definitions) and policy-adherent generation (from API schemas and policy documents). 
Turnstile-trained SLMs (0.6B-4B) match or surpass zero-shot 32B models on multi-turn agentic tasks, and a fine-tuned 0.6B achieves 75.9\% on BFCL (approaching Qwen3-4B at 79.9\%), showing that data quality can compensate for model capacity. 
We also find CoT reasoning is task-dependent: it hurts single-turn function calling through irrelevance rationalization and parameter overthinking, but is critical for multi-turn workflows requiring sequential diagnosis. 
We open-source the framework and release a synthetic dataset with 1K+ APIs and 100K+ interactions.

\paragraphbf{Broader Impact.}
Turnstile enables on-premise generation with open-weight models, avoiding costs and data sovereignty concerns of proprietary APIs, helping democratize synthetic data generation. 
However, as with all synthetic data pipelines, the data may reflect biases present in the teacher model or contain harmful tool-use patterns.  
Our per-step validation and template structure provide natural insertion points for responsible AI guardrails (e.g., content filters, bias checks) at generation time rather than post-hoc.

\paragraphbf{Limitations and Future Work.}
The quality of generated data depends on the teacher LLM capability; stronger models will produce better data. 
Template design currently requires domain expertise, which we aim to address through semi-automated template construction. 
In future work, we plan to extend Turnstile to generate preference pairs for reinforcement learning and validate cross-family transfer with SFT on non-Qwen models.

\bibliographystyle{unsrt}
\bibliography{sections/refs}

\newpage
\appendix

\section{Think vs No-Think: Qualitative Analysis}
\label{sec:appendix:think}
We present qualitative examples illustrating why CoT reasoning hurts single-turn performance but helps multi-turn agentic tasks. We identify three single-turn failure modes (irrelevance rationalization, parameter overthinking, format drift) and shows that multi-turn tasks benefit from structured reasoning. Detailed examples in Tables~\ref{tab:appendix:singleturn_irrel}, ~\ref{tab:appendix:singleturn_overthink} and  Table~\ref{tab:appendix:multiturn_think}.

% ----- Single-Turn Examples -----
\begin{table*}[h]
\centering
\caption{Single-turn: Irrelevance rationalization. Think model reasons itself into calling an inapplicable API; no-think model correctly refuses.}
\label{tab:appendix:singleturn_irrel}
\small
\begin{tabular}{p{0.1\linewidth}|p{0.84\linewidth}}
\toprule
User Request & ``Can you order me a large pepperoni pizza from McDonald's?'' \\
\midrule
Available API & \texttt{uber\_eats.order(restaurant, items, delivery\_address)} \\
\midrule
Think Model & \textit{Reasoning:} ``The user wants to order food. I have uber\_eats.order available. McDonald's does serve food items and could potentially be on Uber Eats. I should try placing the order\ldots'' \newline \textbf{Output:} \texttt{uber\_eats.order(``McDonald's'', [``large pepperoni pizza''], \ldots)} \newline \textbf{Verdict: Incorrect} (should refuse) \\
\midrule
No-Think Model & \textbf{Output:} ``I'm sorry, I cannot help with this. McDonald's does not sell pizza, and I don't have an appropriate tool to order from them.'' \newline \textbf{Verdict: Correct} (refuses appropriately) \\
\bottomrule
\end{tabular}
\end{table*}

\begin{table*}[h]
\centering
\caption{Single-turn: Parameter overthinking on a parallel-multiple call. Think model over-interprets the user request and introduces a spurious constraint; no-think model maps parameters directly.}
\label{tab:appendix:singleturn_overthink}
\small
\begin{tabular}{p{0.1\linewidth}|p{0.84\linewidth}}
\toprule
User Request & ``Book me a one-way flight from SFO to JFK on June 15, and a hotel near Times Square for 2 nights starting June 15.'' \\
\midrule
Available APIs & \texttt{flights.search(origin, dest, date, type)},  \texttt{hotels.book(location, checkin, nights)} \\
\midrule
Think Model & \textit{Reasoning:} ``The user wants a flight and hotel. The flight is one-way SFO to JFK on June 15. For the hotel, they said `near Times Square' --- but since they're flying into JFK, maybe they actually mean near JFK airport for the first night and then Times Square? Let me book for the area near JFK to be safe\ldots'' \newline \textbf{Output:} \texttt{flights.search(``SFO'', ``JFK'', ``2025-06-15'', ``one-way'')},  \texttt{hotels.book(``JFK Airport'', ``2025-06-15'', 2)} \newline \textbf{Verdict: Incorrect} (location should be ``Times Square'') \\
\midrule
No-Think Model & \textbf{Output:} \texttt{flights.search(``SFO'', ``JFK'', ``2025-06-15'', ``one-way'')}, \texttt{hotels.book(``Times Square'', ``2025-06-15'', 2)} \newline \textbf{Verdict: Correct} (maps parameters directly from request) \\
\bottomrule
\end{tabular}
\end{table*}
\newpage

% ----- Multi-Turn Example -----
\begin{table}[H]
\centering
\caption{Multi-turn $\tau^2$-bench Telecom example (task: \texttt{data\_saver\_on|data\_usage\_exceeded}). Customer C1001 has 3 lines; user's line is L1002 (phone 555-123-2002). Think model succeeds via structured diagnosis; no-think model fails by latching onto an irrelevant line.}
\label{tab:appendix:multiturn_think}
\small
\begin{tabular}{p{0.03\textwidth}|p{0.38\textwidth}|p{0.33\textwidth}|p{0.20\textwidth}}
\toprule
& \textbf{Think Model (1.7B)} & \textbf{No-Think Model (1.7B)} & \textbf{Diagnosis} \\
& \footnotesize{42 messages, \textbf{Success}} & \footnotesize{14 messages, \textbf{Failure}} & \\
\midrule
User & \multicolumn{2}{l|}{``My mobile data either stops working or is very slow. I need excellent internet speed.''} & \\
\midrule
Auth & \textit{``I need to authenticate the user. I should ask for their phone number\ldots''} \newline \texttt{\seqsplit{get\_customer\_by\_phone(555-123-2002)} $\to$ C1001} & \texttt{get\_customer\_by\_phone(555-123-2002)} $\to$ C1001 & Both authenticate correctly \\
\midrule
Lines & \textit{``L1002's phone matches the user's number, confirming this is the correct line\ldots''} \newline \texttt{get\_details\_by\_id(L1001)} $\to$ phone 2001 \newline \texttt{get\_details\_by\_id(L1002)} $\to$ phone \textbf{2002} \newline \textit{Stops here.} & \texttt{get\_details\_by\_id(L1001)} $\to$ phone 2001 \newline \texttt{get\_details\_by\_id(L1002)} $\to$ phone 2002 \newline \texttt{get\_details\_by\_id(L1003)} $\to$ \textbf{Suspended} & Think reasons L1002 matches user's phone and stops. No-think blindly iterates all lines. \\
\midrule
Next & \textit{``Correct line identified. I should guide through initial device checks\ldots''} $\to$ proceeds to diagnose L1002 & ``Your line L1003 is suspended due to a contract issue. I can transfer you to a specialist\ldots'' $\to$ user exits (out of scope) & Think focuses on correct line. No-think sees ``Suspended'' on L1003 and assumes this explains the issue. \\
\midrule
Diag. & \textit{``Data Saver is on, this could be causing slowness\ldots''} \newline \texttt{check\_status\_bar()} $\to$ \textbf{Data Saver ON}  \newline \texttt{toggle\_data\_saver\_mode()} $\to$ OFF & --- & Hypothesis formation: thinks about possible cause before acting \\
\midrule
Diag. & \textit{``Still failing after Data Saver fix. Data might be capped\ldots''} \newline \texttt{run\_speed\_test()} $\to$ No Connection \newline \texttt{get\_data\_usage(L1002)} $\to$ 15.1/15.0 GB & --- & Evidence-based pivot: first fix insufficient, reasons about next hypothesis \\
\midrule
Fix & \textit{``Data limit exceeded, I need to refuel\ldots''} \newline \texttt{refuel\_data(C1001, L1002, 2.0)} $\to$ +2 GB \newline \texttt{run\_speed\_test()} $\to$ 220 Mbps & --- & Multi-issue discovery: resolves both root causes sequentially \\
\bottomrule
\end{tabular}
\end{table}

\section{Diversity Metrics: Formal Definitions}
\label{sec:appendix:diversity}

We define four metrics used in Table~\ref{tab:diversity}. Let $\mathcal{D}$ denote a dataset of interactions.

\paragraphbf{(a) Tool Balance.}
Let $c_i$ be the number of calls to tool $i$ across all interactions, and $K$ the number of unique tools. Tool Balance is:
\[
D_{\text{tool}} = \frac{-\sum_{i=1}^{K} p_i \log_2 p_i}{\log_2 K}, \quad \text{where } p_i = \frac{c_i}{\sum_j c_j}
\]
Score of 1.0 means all tools are called equally often; lower values indicate skewed usage.

\paragraphbf{(b) Call Sequence.}
Each interaction is represented as a sequence of tool-call steps. Consecutive API calls without an intervening observation are grouped as a parallel batch (sorted alphabetically). Let $s_j$ be the canonical tool-call sequence for interaction $j$, and $K$ the number of unique sequences. Call Sequence is the normalized entropy over the frequency distribution of these sequences:
\[
D_{\text{seq}} = \frac{-\sum_{k=1}^{K} q_k \log_2 q_k}{\log_2 K}, \quad \text{where } q_k = \frac{|\{j : s_j = k\}|}{|\mathcal{D}|}
\]
Score of 1.0 means every interaction has a unique tool-call ordering.

\paragraphbf{(c) Structure.}
Extract the role sequence for each interaction (excluding SYSTEM), e.g., [USER, THINK, API\_CALL, API\_OBS, ASST]. Compute all 4-grams across all interactions. Structure is the normalized entropy over the 4-gram frequency distribution:
\[
D_{\text{struct}} = \frac{-\sum_{k=1}^{K_4} r_k \log_2 r_k}{\log_2 K_4}
\]
where $r_k$ is the frequency of the $k$-th unique 4-gram and $K_4$ is the number of unique 4-grams. High values indicate diverse interaction structures; low values indicate repetitive patterns.

\paragraphbf{(d) Arg Richness.}
For each tool $t$ with at least $n$ calls, sample $n$ calls uniformly without replacement. Canonicalize each call's arguments as a sorted tuple of (key, lowercase-value) pairs. Compute uniqueness as the fraction of distinct canonical tuples in the sample. Repeat $R$ times and average. Macro-average across all qualifying tools:
\[
\text{ArgRich} = \frac{1}{|T|} \sum_{t \in T} \frac{1}{R} \sum_{r=1}^{R} \frac{|\text{unique}(\text{sample}_r(t, n))|}{n}
\]
where $T = \{t : |\text{calls}(t)| \geq n\}$. We use $n{=}10$, $R{=}50$. A score of 1.0 means every sampled call to a tool has a unique set of argument values; lower values indicate repetitive parameterization.

\section{Ablation Details: Single-Shot Generation}
\label{sec:appendix:single-shot}
To isolate the contribution of role-wise decomposition, we implement a single-shot generation baseline that produces an entire multi-turn interaction in one LLM call using the same teacher model (Qwen2.5-32B-Instruct) and the same scenario distribution for $\tau^2$-bench.

\paragraphbf{Generation Details}
The single-shot prompt provides the LLM with: (1) role format rules (identical to role-wise), (2) all 14 backend API definitions, (3) the backend vs.\ device-side tool distinction, and (4) a \emph{role sequence outline} derived from the sampled scenario---listing the expected sequence of roles (e.g., \texttt{USER: complaint $\to$ THINKING $\to$ API\_CALL: get\_customer\_by\_phone $\to$ API\_OBS $\to$ \ldots $\to$ ASST: resolution}).
This outline mirrors the structure that the role-wise template enforces, giving the single-shot baseline equivalent structural guidance.
Scenario-specific data (customer identifiers, expected API observation values, device report values) is provided separately, with instructions to use observation values only in \texttt{API\_OBS} roles and to respect information asymmetry and causality between user and assistant roles.

We generated $\sim$35k interactions (matching the role-wise volume) using the same vLLM server configuration and generation parameters.
We apply the same structural validators used for role-wise data: JSON parsing, role sequence validity, API call syntax and schema checking. 
No retries are performed, interactions either pass or are discarded.
We observed a 89.4\% pass rate for the structural validations. 
The primary failure modes are invalid API observations (non-JSON or empty objects), missing THINKING before action roles, and truncated interactions. This compares to 96.4\% for role-wise generation with retries (88.9\% without retries).

\paragraphbf{Evaluation and Quality Checks.}
We trained the 1.7B and 4B models on the single-shot generated data - both models fared very poorly on the Telecom benchmark, achieving just 3.5\% and 3.7\% respectively. 
An analysis of the errors revealed two catastrophic failure modes: (i) failed authentication and diagnosis workflows due to hallucinated API parameters, and (ii) model gets stuck in irrecoverable action-reaction loops.

Manual quality review of the training data confirms that structural validity does not imply semantic correctness. 
In a sample of 10 structurally-valid interactions, we find: (1) oracle information leakage in 60\% of cases: the assistant uses customer identifiers before any API call reveals them, (2) false success declarations where the claimed fix would not satisfy the benchmark's environment predicate, and (3) shallow reasoning traces that narrate actions rather than demonstrating decision-making. 
These defects are structurally undetectable but directly harmful to downstream training.
Turnstile helps minimize such pitfalls by: (1) role-wise generation provides a tractable task to the LLM, with focused constraints, (2) the validate-before-generate methodology helps detect hallucinations early and prevent cascading errors.

\end{document}